\RequirePackage{rotating}
\documentclass[isre,nonblindrev]{informs3_pom}
\OneAndAHalfSpacedXI

\usepackage[draft]{hyperref}% -> Set to draft mode before submission
\RequirePackage{fix-cm} 
\usepackage{xspace}

	\usepackage{siunitx}
    \def\sym#1{\ifmmode^{#1}\else\(^{#1}\)\fi}
    \DeclareSIUnit\eur{\officialeuro}
    \DeclareSIUnit\M{M}
    \DeclareSIUnit\k{k}
	
	\usepackage[%
  sort&compress
]{cleveref}
  \crefname{chapter}{section}{sections}
	\Crefname{chapter}{Section}{Sections}

\usepackage{etoolbox}
\robustify\bfseries

\usepackage{isomath}
\usepackage{bm}
\usepackage{enumitem}
\usepackage{colortbl}
\usepackage{tabularx}
\usepackage{longtable}
\usepackage{caption}
\usepackage{booktabs}
\usepackage[flushleft]{threeparttable}
\usepackage{multirow}
\usepackage{lscape}
\usepackage{collcell}
\usepackage{makecell}
\usepackage{svg}

\usepackage{adjustbox} 
\usepackage{listings} 
\usepackage{dcolumn}
\lstdefinestyle{promptbox}{
  basicstyle=\ttfamily\small,
  breaklines=true,
  frame=single,
  columns=fullflexible,
  showstringspaces=false,
  keepspaces=true,
  tabsize=2,
  numbersep=6pt
}
\usepackage{algorithm}

\usepackage{amssymb}
\usepackage{xcolor}
\newcommand{\cmark}{\textcolor{green!60!black}{\checkmark}}
\newcommand{\xmark}{\textcolor{red!70!black}{\textbf{\texttimes}}}

\usepackage{algpseudocode} % for algorithmicx pseudocode

\usepackage{etoolbox}           % <--
\renewrobustcmd{\bfseries}{\fontseries{b}\selectfont}
\renewrobustcmd{\boldmath}{}
\newrobustcmd{\B}{\bfseries}

\usepackage{float}
\usepackage{rotating}

\usepackage{array}
\newcolumntype{L}[1]{>{\raggedright\let\newline\\\arraybackslash\hspace{0pt}}p{#1}}
\newcolumntype{C}[1]{>{\centering\let\newline\\\arraybackslash\hspace{0pt}}p{#1}}
\newcolumntype{R}[1]{>{\raggedleft\let\newline\\\arraybackslash\hspace{0pt}}p{#1}}
\usepackage{subfig}
\usepackage{csquotes}
\usepackage{amsmath}

\usepackage{amssymb}% for \mathbb
\usepackage{pifont}% http://ctan.org/pkg/pifont

\newcommand{\mcellt}[2][c]{%
  \begin{tabular}[t]{@{}#1@{}}#2\end{tabular}}

\usepackage[
    colorinlistoftodos,
    textsize=footnotesize,
        ]{todonotes}

\definecolor{darkgreen}{rgb}{0.0, 0.5, 0.0}

\makeatletter
    \renewcommand{\fps@figure}{H}         % default {tbp}
    \renewcommand{\fps@table}{H}         % default {tbp}
\makeatother 

\usepackage{float}
\usepackage{rotating}
\usepackage{eurosym}

\makeatletter
\newcolumntype{B}[3]{>{\boldmath\DC@{#1}{#2}{#3}}c<{\DC@end}}
\newcolumntype{H}{>{\setbox0=\hbox\bgroup}c<{\egroup}@{}}
\makeatother

\usepackage{natbib}
 \bibpunct[, ]{(}{)}{,}{a}{}{,}%
 \def\bibfont{\small}%
\TheoremsNumberedThrough     % Preferred (Theorem 1, Lemma 1, Theorem 2)
\EquationsNumberedThrough    % Default: (1), (2), ...
\usepackage{eqparbox}

\begin{document}

%%%%%%%%%%%%%%%%
%\renewcommand\thefigure{\arabic{figure}.}
%\setcounter{page}{1}
%\DOI{10.1287/xxxx.0000.0000}%

\RUNAUTHOR{Maarouf, Bakiaj, and Feuerriegel}


\RUNTITLE{Predicting Startup Success Using Large Language Models: A Novel In-Context Learning Approach}

\TITLE{Predicting Startup Success Using Large Language Models: A Novel In-Context Learning Approach}

\ARTICLEAUTHORS{%
\AUTHOR{Abdurahman Maarouf\textsuperscript{\dag}}
\AFF{Munich Center for Machine Learning (MCML) \& LMU Munich, \EMAIL{a.maarouf@lmu.de}}
\AUTHOR{Alket Bakiaj\textsuperscript{\dag}}
\AFF{LMU Munich, \EMAIL{a.bakiaj@campus.lmu.de}}
\AUTHOR{Stefan Feuerriegel}
\AFF{Munich Center for Machine Learning (MCML) \& LMU Munich, \EMAIL{feuerriegel@lmu.de}}

} % end of the block

\begingroup
\renewcommand{\thefootnote}{\dag}
\footnotetext{These authors contributed equally.}
\endgroup

\ABSTRACT{ 
Venture capital (VC) investments in early-stage startups that end up being successful can yield high returns. However, predicting early-stage startup success remains challenging due to data scarcity (e.g., many VC firms have information about only a few dozen of early-stage startups and whether they were successful). This limits the effectiveness of traditional machine learning methods that rely on large labeled datasets for model training. To address this challenge, we propose an in-context learning framework for startup success prediction using large language models (LLMs) that requires no model training and leverages only a small set of labeled startups as demonstration examples. Specifically, we propose a novel $k$-nearest-neighbor-based in-context learning framework, called $k$NN-ICL, which selects the most relevant past startups as examples based on similarity. Using real-world profiles from Crunchbase, we find that the $k$NN-ICL approach achieves higher prediction accuracy than supervised machine learning baselines and vanilla in-context learning. Further, we study how performance varies with the number of in-context examples and find that a high balanced accuracy can be achieved with as few as 50 examples. Together, we demonstrate that in-context learning can serve as a decision-making tool for VC firms operating in data-scarce environments.

}


\maketitle
\sloppy
\raggedbottom
\interfootnotelinepenalty=10000

% NOTES
% - Gao.2024

\section{Introduction}
\label{sec:introduction}

%startups
Startups are newly formed ventures that operate in rapidly changing and highly uncertain environments \citep{Bortolini.2018,Spender.2017}. Only a small share of startups eventually reach successful exits, such as acquisitions or initial public offerings (IPOs), while the vast majority fail, with empirical studies reporting failure rates of around 90\%  \citep{Investopedia.2024}. Despite this high failure rate, investing in early-stage startups is particularly attractive for venture capital (VC) companies and angel investors, since these investments can yield high returns if the startup ultimately succeeds \citep{Buchner.2017}.

%machine learning

Recent research has applied machine learning (ML) methods to predicting startup success \citep[e.g.,][]{Kim.2023,Krishna.2016,RazaghzadehBidgoli.2024,Ross.2021,Thirupathi.2022,Zbikowski.2021}. These methods typically rely on supervised ML techniques and therefore require large labeled datasets. However, many investors and VC firms have historical data from only a few dozen startups \citep{Cumming.2025}. In practice, individual VC firms receive hundreds of potential startup deals per year, but only a small subset progresses through screening and due diligence. For example, one overview of the VC deal flow process describes a typical firm reviewing roughly 200 companies annually, of which about 25 may be advanced to deeper evaluation, only a handful enter formal due diligence, and roughly one to two deals close on average each year \citep{Cumming.2025,Excedr.2024,Sourcescrub.2024}. 
Such limited sample sizes make it difficult to train supervised ML models reliably. With so few labeled examples---often combined with largely unstructured information---models are prone to overfitting and do not generalize well to new startups \citep{Dellermann.2017,Yin.2021}. As a result, predictions based on traditional supervised ML methods are often unstable or unreliable. The central challenge for early-stage startup prediction is therefore learning effectively from extremely small datasets, a setting for which conventional supervised ML methods are poorly suited.

To address the above challenge, we propose a novel few-shot learning framework for early-stage startup success prediction. In this setting, few-shot means that predictions are made using only a small number of labeled startup examples, typically a few dozen available at inference time, rather than by training a model on large labeled datasets. Our framework thus builds on in-context learning (ICL) with pretrained large language models (LLMs) \citep{Brown.2020}, where the prediction task is specified through a prompt containing a small set of example startups together with their observed outcomes. This approach avoids model training or fine-tuning and instead exploits the ability of LLMs to reason by analogy across examples. Although early-stage startups provide only limited and largely textual information, LLMs have been trained on vast and diverse corpora, enabling them to interpret such inputs and draw meaningful comparisons \citep{Brown.2020,Wang.2023}.

In our paper, we operationalize few-shot learning through a $k$-nearest-neighbor-based ICL framework, which we refer to $k$NN-ICL. Our $k$NN-ICL method retrieves a set of startups that are most similar to the target startup and inserts them directly into the prompt as in-context examples. Here, similarity is computed using both structured attributes (e.g., founding year, number of founders) and unstructured information (e.g., short textual self-descriptions), reflecting the heterogeneous data typically available in early-stage investment decisions. By providing the LLM with a targeted set of comparable startups, kNN-ICL creates a contextual ``micro-training set'' at inference time, which eventually enables effective generalization from only a few examples. As a result, our proposed framework is particularly well-suited to data-scarce environments: it requires no model training or fine-tuning, adapts naturally to small datasets, and integrates structured and unstructured data in ways that traditional supervised ML methods cannot.

We evaluate our $k$NN-ICL framework in a setting designed to mimic real-world early-stage VC decision-making, where predictions must be made from only a small number of available startup examples. In each experiment, our approach uses a small database of 4,034 startups reflecting the limited information typically available to investors and then selects only 10, 30, or 50 shots from the database. We curated the database of 4,034 startups using Crunchbase. We chose Crunchbase primarily to ensure reproducibility as well as comparability to earlier work. In general, Crunchbase often lists more mature startups after seed and pre-seed funding and thus offers more extensive information than investors usually observe. Thus, we mimic the setting of VC decision-makers aiming to invest in early-stage startups by limiting the available information for each startup. Across all experiments, our $k$NN-ICL framework outperforms both the supervised ML baselines and vanilla ICL (where the in-context examples are randomly selected from the corpus without our $k$NN approach). The performance increases as more examples are provided, with the best results obtained for 50 shots. In this case, $k$NN-ICL reaches a balanced accuracy of 71.3\%, exceeding the supervised ML baselines (63.1\%) and vanilla ICL (69.6\%). Our results remain robust across different numbers of shots, additional baselines, and sector-level evaluations.

Our work makes three main contributions. First, we contribute to the growing interest in leveraging LLMs for operations research and business analytics \citep[e.g.,][]{ Maarouf.2025, Wu.2025}. However, many of the previous approaches inherit the data-hungry assumptions of traditional supervised ML, which limit their applicability in operational settings where labeled data is scarce. Our work demonstrates that ICL, as an emerging paradigm in ML \citep{Brown.2020}, can overcome these constraints. Here, we thus add by studying the operational value in business decision-making. Second, we show that the choice of examples used for ICL is a crucial determinant of performance. Previously, many ICL applications relied on prompts with a fixed, manually chosen set of examples \citep[e.g.,][]{Toetzke.2022}, meaning that the external knowledge does not adapt to the specific prediction task. Our framework addresses this limitation by selecting examples in a data-driven manner, ensuring that the most relevant instances are included in the prompt. We demonstrate that our $k$-nearest-neighbor-based retrieval approach substantially improves prediction performance. This finding has important managerial implications: organizations should dynamically curate example sets rather than rely on static prompt templates. Third, our $k$NN-ICL approach is widely applicable to data-scarce operational prediction tasks. We see promising applications in, for example, business failure forecasting \citep[e.g.,][]{Borchert.2023}, credit risk evaluation \citep[e.g.,][]{Gunnarsson.2021,Kriebel.2022}, and other domains in which only limited structured data is available.

%outline
The rest of this work is structured as follows. \Cref{sec:related_work} reviews prior work on startup success prediction and related applications of natural language processing in operations research and business analytics. In \Cref{sec:method}, we introduce our $k$NN-ICL framework and detail how it supports early-stage prediction of startup success. \Cref{sec:experiment_setup} specifies our experimental setup, including the data, baselines, and evaluation metrics. We then evaluate the prediction performance (\Cref{sec:results}), before discussing implications for research and practice (\Cref{sec:disscusion}). \Cref{sec:conclusion} concludes.

\section{Related Work}
\label{sec:related_work}

\subsection{Venture Capital}
\label{sec:VC}

Startups operate under high uncertainty, and only a minority achieve growth and, ultimately, successful exits. Identifying promising ventures early is therefore a central challenge in VC decision-making  \citep{Gompers.2020}. Prior research has examined which observable characteristics correlate with later success, including different features of the business model \citep{Bohm.2017,Weking.2019}, founder background and experience \citep{Ratzinger.2018}, and early financing activity, which can function as quality signals \citep{Nahata.2008}. Yet, the VC selection process often remains heuristic. Common approaches from VC practice, including traditional scorecards and manual screening methods \citep{Dellermann.2017,Huang.2015}, both of which have inherent limitations due to the high degree of subjectivity. Below, we therefore review data-driven approaches.

\subsection{NLP In Business Analytics}
\label{sec:nlp_business_analytics}

Natural language processing (NLP) is nowadays widely used in business analytics, especially for extracting information from unstructured text data such as product reviews \citep[e.g.,][]{Archak.2011,Netzer.2012,Tirunillai.2014}, social media \citep[e.g.,][]{Ducci.2020}, financial disclosures  \citep[e.g.,][]{Kraus.2017}, or websites \citep[e.g.,][]{Borchert.2023}. Early approaches primarily relied on bag-of-words or term-frequency representations combined with traditional ML classifiers such as logistic regression or support vector machines \citep{Feuerriegel.2025}. Recently, transformer-based models and language models such as BERT \citep{Devlin.2019} have substantially improved the ability to extract and leverage information from text data for forecasting and decision support. These advances have contributed to the growing interest in leveraging deep learning such as LLMs for operations research and business analytics \citep{Fischer.2018,Kraus.2020,Kriebel.2022,Maarouf.2025,Stevenson.2021}. However, these methods typically require large labeled datasets for fine-tuning the underlying ML classifier, which limits the applicability in many business settings, such as early-stage VC, where only a small number of labels are available.

In contrast, ICL offers a practical remedy by leveraging LLMs to make predictions directly from examples provided at inference time in a zero-shot or few-shot setting. Here, zero-shot learning refers to generating predictions without any task-specific labeled data, while few-shot learning provides the model with a small number of labeled examples to make predictions. The key enabler behind zero-shot and few-shot inferences is nowadays pre-trained LLMs, which encode broad conceptual knowledge. ICL operationalizes these abilities by placing a small set of examples in the prompt and asking the model to generalize to a new instance. This makes ICL well-suited for data-scarce environments, because it avoids the need for large labeled datasets and parameter fine-tuning. However, the performance depends on which examples are selected for the prompt \citep{Peng.2024}. Prior work often relies on manually chosen examples \citep[e.g.,][]{Toetzke.2022}, which may not correspond to cases with the best predictive power. Moreover, ICL is subject to a trade-off \citep[e.g.,][]{Rubin.2022,Schulhoff.2025}: while providing more examples can improve performance, including irrelevant or weakly related examples may hamper generalization. To address this challenge, we later introduce a $k$-nearest-neighbor retrieval approach that automatically selects a subset of relevant examples for ICL in a data-driven manner.

\subsection{Predicting Startup Success}
\label{sec:relatedwork_predicting_startup_success}

% ML & DL success prediction based on large-scale data

One stream of research has applied supervised ML models to predict startup success \citep[e.g.,][]{Choi.2024,Wei.2025} Early work developed classifiers that use structured features about firm characteristics, such as founding year, funding history, or team composition, to estimate the likelihood of future growth or exit \citep{RazaghzadehBidgoli.2024}.. However, such supervised ML models are trained to capture complex patterns, but such training requires large labeled datasets that are typically unavailable for early-stage startups. As a result, existing supervised ML approaches are difficult to apply when VCs must make predictions from a small-scale dataset with only a few dozen labeled cases. Further, existing supervised ML approaches are typically confined to leveraging structured features, thus leaving the predictive power of textual features untapped.

In principle, investors could draw on public startup databases such as Crunchbase or PitchBook. However, these platforms are designed as broad, standardized repositories and therefore capture only a limited subset of the information that investors typically rely on when evaluating early-stage startups. In particular, VC firms often possess proprietary data that is not available in public databases, such as internal market assessments, evaluations from expert networks, detailed feedback from pitch meetings, or insights generated during due diligence. Moreover, information in public databases is often accumulated only after startups reach more advanced stages, leading to self-selection \citep{Cader.2011}, which limits the usefulness during seed and pre-seed investment decisions. As a result, investors have strong incentives to rely on and analyze their own internally collected data rather than depend solely on public datasets.

Another line of research integrates structured and unstructured data to predict startup success. Textual descriptions can capture information that is difficult to encode in structured form, such as a startup’s business model, strategic focus, or target market---and thus provide valuable predictive signals \citep[e.g.,][]{Gavrilenko.2023,Maarouf.2025,Park.2024,Potanin.2023,Te.2023}. A few recent studies use zero-shot prompting to leverage LLMs directly from text, typically by asking the model to classify startups or generate rationales without relying on historical labeled data \citep{Griffin.2025,Mu.2025}. While these works illustrate the potential of LLMs to reason over startup narratives, they do not make use of past startup outcomes and therefore do not fully exploit available historical information. One study uses fine-tuning to incorporate textual data \citet{Maarouf.2025} by computing text embeddings using an LLM and then fine-tuning a supervised prediction model (e.g., a linear regression, neural network) on top of these embeddings, but where fine-tuning the underlying prediction model still requires a large labeled dataset as input.

The approaches above rely either on fine-tuning models with large labeled datasets or on zero-shot prompting that ignores historical information (see \Cref{tab:prior_studies}). In contrast, early-stage VC settings require methods that can learn effectively from only a small number of labeled examples. To the best of our knowledge, no existing work provides a few-shot framework for startup prediction that leverages past examples about startups without requiring model training. We address this gap by using ICL; however, as we later show, standard ICL with a randomly chosen set of examples performs poorly. As a solution, we therefore develop a novel retrieval approach, called $k$NN-ICL.

\begin{table}[htbp]
\centering
\caption{Overview of key studies on startup success prediction.}
\label{tab:prior_studies}
\small
\begin{adjustbox}{max width=\textwidth}
\begin{tabular}{@{}>{\raggedright\arraybackslash}p{4.0cm}
                >{\raggedright\arraybackslash}p{6.2cm}
                >{\centering\arraybackslash}p{2.6cm}
                >{\centering\arraybackslash}p{3.0cm}
                >{\raggedright\arraybackslash}p{4.2cm}@{}}
\toprule
\textbf{Approach} &
\textbf{Key studies} &
\textbf{Features used} &
\textbf{Inference-time only (=no training)} &
\textbf{Suitable for small labeled datasets} \\
\midrule

Traditional supervised ML &
\cite{Choi.2024, Park.2024, Potanin.2023, RazaghzadehBidgoli.2024, Te.2023, Yin.2021} &
Structured &
\xmark &
\xmark \\

\midrule

Zero-shot LLM  &
\cite{Griffin.2025,Mu.2025} &
Structured &
\cmark &
n/a \textit{(ignores historical outcomes)} \\

\midrule

Fine-tuned LLM &
\cite{Maarouf.2025} &
Structured + textual &
\xmark &
\xmark  \\

\hline
\hline

Few-shot LLM &
$k$NN-ICL (\textbf{ours}) &
Structured + textual &
\cmark  &
\cmark \\

\bottomrule
\end{tabular}
\end{adjustbox}
\end{table}

\section{Method}
\label{sec:method}

% Our research objective
In this section, we present our few-shot framework for predicting startup success in data-scarce settings. We first formulate the prediction problem of a VC firm aiming to invest in early-stage startups (see \Cref{sec:problem_statement}) and then propose our $k$NN-ICL approach (see \Cref{sec:icl_approach}).

\subsection{Problem Formulation}
\label{sec:problem_statement}

We study the setting in which a VC firm aims to predict the future success or failure of early-stage startups to support investment decision-making. The task of predicting the future success of early-stage startups is challenging due to data scarcity. A typical VC firm has evaluated only a relatively small number of startups with known outcomes, often only a few dozen cases \citep{Cumming.2025,Excedr.2024,Sourcescrub.2024}. Such limited data make it difficult to train traditional supervised ML models and thus to infer reliable relationships between early indicators and eventual success.

Another challenge is that the information available for early-stage startups consists of a mix of structured and textual features. Structured data, such as founding year, sector, or team size, are known to be linked to startup success \citep[e.g.,][]{Gavrilenko.2023,Te.2023}. Recent literature has also emphasized the value of qualitative information in the form of short textual descriptions, which often convey aspects of the business model, product focus, or market segment \citep[e.g.,][]{Gavrilenko.2023, Maarouf.2025, Sharchilev.2018}. However, training ML models on such a combination of structured and unstructured data typically requires large labeled datasets, which stands in direct contrast to the data-scarce setting that is encountered in VC practice.

Because of the above challenges, traditional approaches based on supervised ML \citep[e.g.,][]{Gonzalez-Gutierrez.2024, Yin.2021} are often not practical. Instead, we propose using ICL to build a few-shot learning framework, in which the model receives only a small number of labeled examples, consistent with what VC firms observe in practice. ICL \citep[e.g.,][]{Brown.2020, Dong.2024} does not require explicit training or fine-tuning; rather, it performs reasoning at inference time using the examples provided in the prompt. However, few-shot ICL introduces a trade-off: including more examples may provide richer information, but too many or poorly matched cases can reduce relevance and thus hinder generalization \citep{Schulhoff.2025}.

In general, the effectiveness of ICL depends heavily on the relevance and composition of the selected examples \citep{Rubin.2022}. In our VC application, randomly choosing a set of historical startups will yield examples that do not resemble the target startup for which the prediction is to be made and can thus reduce overall predictive accuracy. This issue is amplified by the highly imbalanced outcome distribution in early-stage settings, where most startups eventually fail \citep{Park.2024}. If startups are selected without careful selection, the prompt may contain predominantly negative cases, causing the model to reflect the overall failure rate rather than learn meaningful distinctions between successful and unsuccessful startups. To address this challenge, we introduce our $k$NN-ICL approach that learns which examples are most relevant for each prediction and balances the distribution between successful vs. failed startups.

\subsection{Proposed In-Context Learning Approach}
\label{sec:icl_approach}

Let $i = 1, \ldots, n$ denote startup $i$ in the historical dataset $\mathcal{H}$ of the VC firm. In particular, the historical dataset contains the previously evaluated startups, for which outcomes are known.
Typically, $n$ is in the order of a few dozen, reflecting the real-world data constraints in VC practice \citep{Cumming.2025,Excedr.2024,Sourcescrub.2024}. For each historical startup, we consider three types of information: (1)~Structured data (SD) represented as $\mathbf{x}^{\text{SD}}_i \in \mathbb{R}^{m}$ describes company-level characteristics such as company age, founder characteristics, and sector indicators. (2)~Textual data (TD) refers typically to short self-descriptions, which we convert into a dense embedding vector $\mathbf{x}^{\text{TD}}_i \in \mathbb{R}^{d}$ using a pretrained model that captures both semantic and syntactic properties of the text. Note that both variables, $\mathbf{x}^{\text{SD}}_i$ and $\mathbf{x}^{\text{TD}}_i$, are later handled in different ways to account for the structured versus unstructured nature of the variable, respectively. (3)~Each historical startup also has an observed outcome label $y_i \in \{0,1\}$ indicating whether the startup was successful or not within the defined success horizon.  Formally, we write the historical dataset as \begin{equation}
\mathcal{H} = \left\{ \left(\mathbf{x}^{\text{SD}}_i, \mathbf{x}^{\text{TD}}_i, y_i \right) \right\}_{i=1, \ldots, n}.
\end{equation}

Our goal is to make a prediction for a \emph{new} startup, hereafter referred to as the \textit{target startup}. We refer to the new startup as $\tau$. Let its structured variables be denoted by
$\mathbf{x}^{\text{SD}}_{\tau}$ and its textual embedding by $\mathbf{x}^{\text{TD}}_{\tau}$. We aim to predict the target startup’s binary outcome $\tilde{y}_{\tau} \in \{0,1\}$.

\begin{figure}[htbp]
    \centering
    \includegraphics[width=\textwidth]{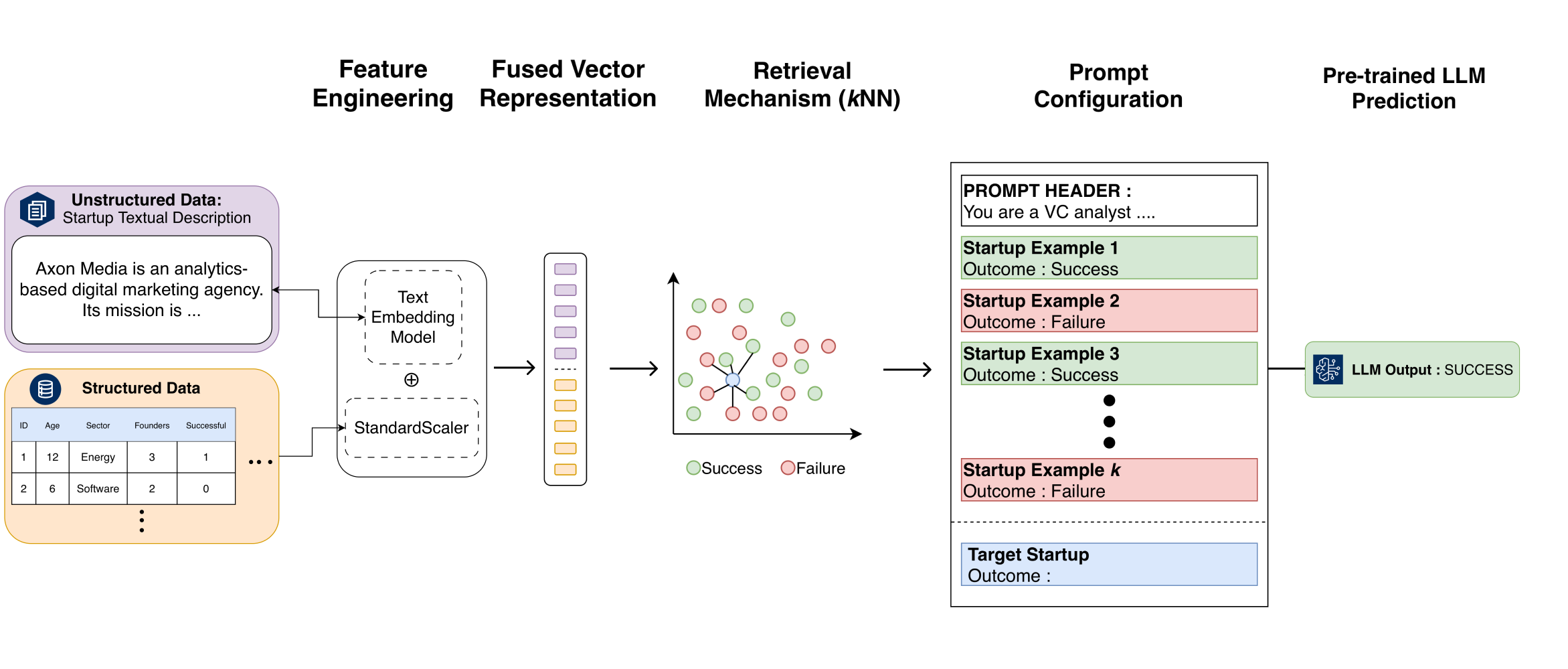}
    \caption{Overview of the proposed $k$NN-ICL prediction pipeline.}
    \label{fig:knn_pipeline}
\end{figure}

Our $k$NN-ICL approach for predicting startup success proceeds as follows (see Figure~\ref{fig:knn_pipeline} for an overview). For each target startup, a set of $k < n$ relevant examples of past startups are retrieved from the training set $\mathcal{H}$ based on similarity with regard to the structured data and textual data. Formally, we retrieve a context set
\begin{equation}
\mathcal{C}_{i} = \left\{ \left(\mathbf{x}^{\text{SD}}_i, \mathbf{x}^{\text{TD}}_i, y_i \right) \right\}_{i=1, \ldots, k} \subset \mathcal{H}.
\end{equation}

In the following, let $\Psi(\cdot)$ denote the LLM\footnote{\SingleSpacedXI\footnotesize
Because LLMs may reproduce memorized information from pretraining data, we take care to prevent information leakage in our evaluation. In particular, we redact startup identifiers by replacing company names in textual self-descriptions with a placeholder (``this company''), so that predictions cannot rely on name-based memorization.} and $\mathcal{P}(\mathcal{C}_\tau)$ is the prompt constructor that assembles the retrieved examples $C_\tau$ into a standardized textual input. The LLM generates a prediction $\tilde{y}_\tau \in \{\texttt{"SUCCESS"}, \texttt{"FAILURE"}\}$ as a text string, which is mapped to a binary output through a decoding function $g(\cdot)$. Crucially, our $k$NN-ICL approach only queries the LLM and thus performs the prediction directly at inference time  (i.e., and thus circumvents the need for fine-tuning). In the following, we describe the retrieval mechanism that identifies the most relevant $k$ startups to include as examples (shots) for each prediction (Section~\ref{sec:retrieval}) and the prompt constructor $\mathcal{P}(\mathcal{C}_\tau)$  (Section~\ref{sec:prompt}).

\subsubsection{Retrieval Mechanism}
\label{sec:retrieval}

In our framework, the retrieval mechanism identifies the support set of most comparable past startups ($C_\tau$) for each target startup ($\tau$) through a two-step process. First, we construct a unified representation for every startup $i$ by combining its structured attributes ($\mathbf{x}^{\text{SD}}_{i} \in \mathbb{R}^{m}$) with the embedding of its textual description ($\mathbf{x}^{\text{TD}}_{i} \in \mathbb{R}^{d}$), yielding a fused vector ($\mathbf{z}_i \in \mathbb{R}^{m+d}$). This representation allows structured and unstructured information to be compared in a single feature space. Second, for a given target startup, we compute its similarity to all past startups using cosine similarity ($s(i,\tau)$), rank the historical firms by this score, and select the $k/2$ most similar \texttt{"SUCCESS"} and $k/2$ most similar \texttt{"FAILURE"} startups. These retrieved examples form the support set $C_\tau$ shown to the LLM during ICL, thereby ensuring that the model reasons from startups most similar to the target startup $\tau$. We detail both steps in the following.

In the first step, we take the textual self-description of each startup and encode it into a dense numerical embedding using a pretrained embedding model. This pretrained model converts free-form text into high-dimensional vectors that encode semantic and syntactic relationships across languages and writing styles. Each textual description is passed to an embedding model, which internally performs tokenization and normalization during encoding. The output is a high-dimensional vector $\mathbf{x}^{\text{TD}}_{i} \in \mathbb{R}^{d}$ that represents the textual description of startup $i$. Alongside the textual description, we incorporate structured data via numerical vectors $\mathbf{x}^{\text{SD}}_{i} \in \mathbb{R}^{m}$. Then, the structured and unstructured data are concatenated, so that we yield the fused vector
\begin{equation}
\mathbf{z}_i = [\,\mathbf{x}^{\text{SD}}_{i};\mathbf{x}^{\text{TD}}_{i}\,] \in \mathbb{R}^{m+d} ,
\end{equation}
where we denote concatenation by the operator $[\,\cdot\,;\,\cdot\,]$. Before concatenation, we standardize the structured features within each train-test split and rescale the structured block so that
\begin{equation}
\lVert \mathbf{x}^{\text{SD}}_{i}\rVert_2 = \alpha\,\lVert \mathbf{x}^{\text{TD}}_{i}\rVert_2,
\end{equation}
where $\alpha$ controls the relative influence of structured versus textual information in cosine similarity retrieval. We set $\alpha=0.5$ to yield comparable contributions of structured and textual features under cosine similarity. We later perform a sensitivity analysis for different values of $\alpha$ in \Cref{sec:scaling}.

In the second step, we rank and select the startups. Specifically, for the target startup $\tau$, we compute the fused representation $\mathbf{z}_\tau$ analogously to all other startups in the dataset. To identify comparable past startups, we measure similarity between $\tau$ and each training startup $i \neq \tau$ using cosine similarity, i.e.,
\begin{equation}
s(i,\tau) = \frac{\mathbf{z}_i^\top \mathbf{z}_\tau}{\lVert \mathbf{z}_i\rVert_2 \,\lVert \mathbf{z}_\tau\rVert_2}.
\end{equation}
All training startups are ranked by $s(i,\tau)$, and the $k$-nearest-neighbors with the highest similarity scores are selected as examples, where $k$ corresponds to the number of examples shown to the model (e.g., $k=$ 10, 20, 30, or 50 shots). To prevent class imbalance, we adopt a stratified approach, where an equal number of \texttt{SUCCESS} and \texttt{FAILURE} examples are drawn from the top-ranked results. This balancing technique corrects the outcome imbalance and ensures both classes are equally represented, rather than letting the LLM be dominated by the majority class. The retrieved startups provide the so-called support set for ICL, which we refer to as $\mathcal{C}_\tau$ and which we provide as context to the LLM to predict the outcome of the target startup.

\paragraph{Implementation details.}
For every target startup, we compute cosine similarity between the fused vector $\mathbf{z}_\tau$  of the target setup and all fused vectors in the training data (the fused vectors are pre-computed per split for numerical efficiency). We then build a strictly class-balanced set with $k$ shots by independently taking the most similar successes and failures (with a small buffer) and oversampling with replacement if a class is short. To reduce positional bias, shots are interleaved, starting from the class whose nearest neighbor is more similar (where near ties are broken at random).

\subsubsection{Prompting}
\label{sec:prompt}

To predict the outcome of a target startup $\tau$, we use the LLM together with the retrieved support set $\mathcal{C}_\tau$ and a tailored prompt. The overall idea is as follows: the LLM receives a small set of labeled examples and then infers the label of the target startup by reasoning over this context. This prompt-based inference allows all learning to occur at prediction time, without any parameter training or model fine-tuning.

In our ICL framework, we follow established best practices in prompt engineering \citep{Dong.2024,Schulhoff.2025,Zhao.2021}. Our prompt consists of four main elements: (i)~a header that defines the task, (ii)~a sequence of labeled examples, (iii)~a final query for which the model must generate a prediction, and (iv)~the demonstrations from the support set. In this study, the prompt is designed to emulate the reasoning process of a VC analyst. Hence, the prompt specifies the model’s role (``You are a VC analyst''), and constrains the response format (``Respond with exactly one word: SUCCESS or FAILURE''). The exact prompt is stated in Algorithm~\ref{alg:icl-prompt}.\footnote{\SingleSpacedXI\footnotesize We also experimented with multiple other prompt formulations but observed only minor variations in performance.} Further, each example in the support set $\mathcal{C}_\tau$ is provided to the LLM with both structured and unstructured data. The support set is added to the prompt into a concise ``\texttt{Startup:}'' field, followed by the observed outcome (i.e., \texttt{"SUCCESS"} or \texttt{"FAILURE"}). The resulting sequence of examples serves as contextual guidance for the LLM when predicting the outcome for the target startup, which is appended at the end of the prompt without its label.

Finally, the assembled prompt $\mathcal{P}(\mathcal{C}_\tau)$ is provided to the LLM $\Psi(\cdot)$, which generates a one-word output following the instructed response format within our prompt; i.e., either \texttt{"SUCCESS"} or \texttt{"FAILURE"}). The textual output is mapped to a binary label through a decoding function $g(\cdot)$, which yields
\begin{equation}
\tilde{y}_\tau = g\!\left(\Psi\!\big(\mathcal{P}(\mathcal{C}_\tau)\big)\right) \in \{0,1\}.
\end{equation}
This approach allows each prediction to be conditioned on the most relevant historical context rather than relying on global parameter training. 

\begin{figure}[!htb]
\captionsetup{type=algorithm}
\centering
\begin{minipage}{\linewidth}
\begin{lstlisting}[
    style=promptbox,
    caption={ICL prompt used in our experiments.},
    label={alg:icl-prompt},
    captionpos=b
]
You are a VC analyst. 
Your task is to predict whether a startup will succeed or not.
SUCCESS if the startup has completed an IPO, received funding, or been acquired.
Otherwise FAILURE.
Each example contains the startup's details and its historical outcome.
Respond with exactly one word: SUCCESS or FAILURE.

[EXAMPLES]
Startup: [Structured + description text for Example 1]
Outcome: SUCCESS

Startup: [Structured + description text for Example 2]
Outcome: FAILURE

...

Startup: [Structured + description text for Example k]
Outcome: FAILURE

===== TARGET STARTUP =====
Startup: [Structured + description text for TARGET startup]
Outcome:
\end{lstlisting}
\end{minipage}
\end{figure}

\section{Experimental Setup}
\label{sec:experiment_setup}

\subsection{Dataset}
\label{sec:trainingcohort}

We curate our evaluation dataset to reflect the information environment faced by VC investors at the time of early-stage screening. In particular, we aim to avoid relying on signals that are only observable for mature startups (e.g., Series C funding or later stages), which would not be available during initial investment decisions. Following the data construction in \citep{Maarouf.2025}, we therefore intentionally assemble an evaluation cohort of early-stage startups that mirrors real-world VC conditions. The cohort consists of 4,034 US-based startups founded between 2013 and 2015 that contain both structured and unstructured information, and exclude companies that have gone public or reached Series C funding or beyond (i.e., limiting the timeframe to this period allows us to observe any event indicative of startup success over a longe time frame between 2016 and 2020). This filtering ensures a focus on early-stage ventures and limits look-ahead bias.

Our sample is drawn from the Crunchbase dataset to maintain reproducibility through the use of public data. Crunchbase is a widely used startup intelligence platform in both academic research and investment practice, which provides company profiles that combine structured attributes and textual descriptions, curated by verified employees in a manner similar to information used in many VC selection processes. While Crunchbase offers broad startup coverage, its raw data tend to overrepresent more mature firms. By applying the above filtering strategy, we construct a dataset that is modest in size, collects startups that are early in stage, and is equipped with the same types of structured and textual signals available to investors at the time of screening, enabling a realistic evaluation of our ML approach. We use data from Crunchbase with a knowledge cut-off of 2015 to ensure that there is no information leakage, so that all data are observable to VC firm at the time of decision-making.

\subsection{Operationalization of Startup Success}
\label{sec:startupsuccess}

In this study, we follow previous research \citep{Arroyo.2019,Hegde.2021} and define startup success using standard outcomes in VC research as follows.  A startup is labeled ``successful'' if the startup experienced at least one major positive event within a five-year window (i.e., between 2016 and 2020): an acquisition, an IPO, or a follow-on funding round. If none of these events occurred, the startup is labeled ``unsuccessful''. Choosing the 2016–2020 window ensures that the outcome labels represent forward-looking developments from the standpoint of an investor making an investment decision at the end of 2015. By isolating outcomes along this clear timeline, the evaluation avoids information leakage that could occur if the model had access to future information at prediction time. To ensure the evaluation represents actual predictive performance, the analysis therefore uses a time-aware design: all input data comprises only information that would have been observable as of the end of 2015, and any data added after 2015, such as later funding rounds, new co-founders, or updated business descriptions, are excluded from the features used to make predictions.

\subsection{Dataset Description}
\label{sec:dataset_description}

Informed by prior research \citep{Kaiser.2020,Maarouf.2025,Sharchilev.2018}, we make use of a rich, high-dimensional set of both structured and unstructured data inputs for prediction. (1)~The structured data capture core characteristics of the startups, including the age at the time of prediction, number of founders, industry classification, and presence on professional platforms (e.g., LinkedIn). The latter is informed by prior research that demonstrates a correlation between social media presence and obtaining funding \citep{Banerji.2019}. Additionally, we include startup founder attributes related to gender diversity, international background, and educational attainment, which are known to be associated with entrepreneurial outcomes \citep{Colombo.2005}. (2)~We further incorporate unstructured textual self-descriptions written by the startups themselves in their Crunchbase profiles. These self-descriptions allow firms to summarize their value proposition, business model, or market strategy. A complete overview of the variables used in retrieval and prediction is presented in Table~\ref{tbl:variables}.

\begin{table}[htb]
\TABLE
{Dataset description.\label{tbl:variables}} 
{
\OneAndAHalfSpacedXI
%\footnotesize
\scriptsize
\begin{tabular}{p{5cm}p{11cm}}
\toprule
\textbf{Variable} & \textbf{Description}  \\
\midrule
			\multicolumn{2}{c}{\textsc{Outcome variable}} \\
			\midrule
  Success & True ($=1)$ if a startup had an initial public offering, received funding, or has been acquired. False ($=0$) otherwise.\\ 
				\midrule
			\multicolumn{2}{c}{\textsc{Predictors}} \\
			\midrule
%\multicolumn{2}{l}{\underline{Company-specific information}} \\

\multicolumn{2}{l}{\underline{Structured Data}} \\
  Age & Time since the startup has been founded (in months) \\ 
  \addlinespace
  Has email & Whether the startup has added an email address ($=1$ if true, otherwise 0) \\ 
  Has LinkedIn & Whether the startup has added a link to LinkedIn ($=1$ if true, otherwise 0) \\ 
  
  \addlinespace
%\multicolumn{2}{l}{\underline{Founders information}} \\
  Founders count & Number of founders of the startup \\ 
  Founders country count & Number of unique countries the founders are from \\ 
  Founders male count & Number of male founders \\ 
  Founders female count & Number of female founders \\ 
  Founders degree count total & Total number of university degrees of the founders \\ 
  
  \addlinespace
%\multicolumn{2}{l}{\underline{Funding information}} \\

	\addlinespace
	Industries & Fine-grained industries in which the startup operates (according to the Crunchbase coding scheme; e.g., ``machine learning'', ``machinery manufacturing'') \\

  \addlinespace
\multicolumn{2}{l}{\underline{Unstructured Data}} \\
  Textual self-description & The startup’s free-text business description. \\

\bottomrule

\end{tabular}
}
{}
\end{table}

\subsection{Dataset Statistics}
\label{sec:dataset_statistics}

Overall, our dataset consists of 4,034 startups. Descriptive statistics on these startups are presented in Table~\ref{tbl:descriptives}. Among them, 1,418 (35.15\%) are classified as successful, while 2,616 (64.85\%) are non-successful startups. In our cohort, startups that list a LinkedIn profile achieve a success rate of 42.3\%, whereas those without one achieve only 17.4\%, indicating a connection between LinkedIn presence and positive outcomes. The average age of startups in our filtered cohort is 18.2 months at the time of observation. Furthermore, startups founded by teams of two or more individuals achieve a success rate of approximately 54\%, compared to just 38\% among single-founder startups. This suggests that having a founding team is positively associated with startup success. Successful startups tend to have slightly more internationally diverse founding teams, with an average of 1.23 founders from different countries, compared to 1.15 among non-successful startups. 

Table~\ref{tab:example_descriptions} provides examples of textual self-descriptions from one successful and one unsuccessful startup.

\begin{table}[H]
\TABLE
{Descriptive statistics.\label{tbl:descriptives}}
{
\OneAndAHalfSpacedXI
%\footnotesize
\scriptsize
\sisetup{round-mode=places,round-precision=2}
%\begin{tabular}{llllll}
\begin{tabular}{lS[table-format=2.2]S[table-format=2.2]S[table-format=2.2]S[table-format=2.2]S[table-format=2.2]S[table-format=2.2]}
%\begin{tabular}{l *{5}{R[1.4cm]}}
  \toprule
\textbf{Variable} & \multicolumn{2}{c}{\textbf{Overall}} & \multicolumn{2}{c}{\textbf{Non-successful}} & \multicolumn{2}{c}{\textbf{Successful}} \\
\cmidrule(lr){2-3} \cmidrule(lr){4-5} \cmidrule(lr){6-7}
& {\textbf{Mean}} & {\textbf{SD}} & {\textbf{Mean}} & {\textbf{SD}} & {\textbf{Mean}} & {\textbf{SD}} \\
% {\textbf{Mean}} & {\textbf{Median}} & {\textbf{Max}} & {\textbf{Min}} & {\textbf{SD}} \\ 
  \midrule
  
			\multicolumn{7}{c}{\textsc{Outcome variable}} \\
			\midrule
  Success & 0.35 & 0.48 & 0.00 & 0.00 & 1.00 & 0.00 \\   
				\midrule
			\multicolumn{7}{c}{\textsc{Predictors}} \\
			\midrule  
%\multicolumn{2}{l}{\underline{Company Specific}} \\
  Age (in months) & 18.23 & 10.08 & 19.57 & 9.93 & 15.77 & 9.88 \\
	\addlinespace
  Has email & 0.77 & 0.42 & 0.75 & 0.43 & 0.81 & 0.40 \\
  Has LinkedIn & 0.71 & 0.45 & 0.64 & 0.48 & 0.86 & 0.35 \\
  \addlinespace
%\multicolumn{2}{l}{\underline{Founders Information}} \\
  Founders count & 1.84 & 0.99 & 1.66 & 0.88 & 2.03 & 1.07 \\
  Founders different country count & 1.19 & 0.42 & 1.15 & 0.38 & 1.23 & 0.45 \\ 
  Founders male count & 1.59 & 1.03 & 1.41 & 0.90 & 1.79 & 1.12 \\
  Founders female count & 0.25 & 0.53 & 0.26 & 0.51 & 0.24 & 0.54 \\
  Founders degree count total & 1.17 & 1.49 & 0.87 & 1.23 & 1.50 & 1.68 \\ 
  
    \midrule
    Textual description length in chars & 670.76 & 427.03 & 704.23 & 466.66 & 609.03 & 333.55 \\

   \bottomrule
   \multicolumn{3}{l}{SD = standard deviation} &  \multicolumn{4}{r}{$N$ = 4,034 startups}
\end{tabular}
}
{
%\hspace{-0.5cm} \emph{\emph{Note:} SD = standard deviation}
}
\end{table}

\begin{table}[htbp]
  \centering
  \caption{Exemplary textual self-descriptions.}
  \label{tab:example_descriptions}
  \scriptsize                           % globally smaller text
  \setlength{\tabcolsep}{4pt}           % tighten column padding
  \begin{adjustbox}{width=\linewidth}   % auto-shrink if still too wide
    \begin{tabular}{p{0.82\linewidth} l}
      \toprule
      \textbf{Textual self-description} & \textbf{Outcome} \\
      \midrule
      \textit{``Axon Media is an analytics-based digital marketing agency in New York City. Its mission is to develop high-quality, integrated strategies that maximise impact. Services span paid, social, content and email marketing plus design and web development. The agency leverages research, data and analytics to optimise campaigns for technology startups and non-profit clients.''} & non-successful \\[0.8ex]
      \midrule
      \textit{``Mosaic collaborates with leading mobility firms—OEMs, ride-sharing services, content and service providers—to build next-generation location-based AI interfaces. The team brings expertise in AI, IoT and location-based services, having worked at Google, Cisco, Apple and other major tech companies. Their products already serve consumers and enterprises at global scale.''} & successful \\
      \bottomrule
    \end{tabular}
  \end{adjustbox}
\end{table}

Startups in our dataset are active across various Global Industry Classification Standard (GICS) sectors\footnote{The Global Industry Classification Standard (GICS) is an industry taxonomy developed by MSCI and Standard \& Poor's and consists of 11 sectors (e.g., \textsc{Energy}, \textsc{Materials}).} (Table~\ref{tbl:sector_allocation}), with the highest representation in \textsc{Information Technology} (55.4\%) and \textsc{Communication Service}s (47.9\%). Less common sectors include \textsc{Utilities} (2.2\%) and \textsc{Materials} (1.5\%). Success rates vary considerably: \textsc{Utilities} (52.2\%) and \textsc{Healthcare} (49.8\%) have the highest, while \textsc{Communication Services} accounts for 32.5\%. Note that startups may belong to multiple sectors.

\begin{table}[H]
\TABLE
{Relative frequencies and success rates of startups across different business sectors.\label{tbl:sector_allocation}}
{
\OneAndAHalfSpacedXI
%\footnotesize
\scriptsize
\sisetup{round-mode=places,round-precision=2}
%\begin{tabular}{llllll}
\begin{tabular}{lS[table-format=2.2]S[table-format=2.2]}
%\begin{tabular}{l *{5}{R[1.4cm]}}
    \toprule
  {\textbf{Business sector}} & {\bfseries\mcellt{Relative\\ freq. (in\,\%)}} & {\bfseries\mcellt{Success rate\\ (in\,\%)}} \\
  \midrule
  \textsc{Information Technology} & 55.4 & 39.9 \\ 
  \textsc{Communication Services} & 47.9 & 32.5 \\ 
  \textsc{Consumer Discretionary} & 30.4 & 34.0 \\ 
  \textsc{Industrials}            & 29.7 & 45.8 \\ 
  \textsc{Health Care}            & 16.4 & 49.8 \\ 
  \textsc{Financials}             & 9.5  & 38.0 \\ 
  \textsc{Consumer Staples}       & 8.3  & 41.7 \\ 
  \textsc{Real Estate}            & 6.0  & 35.1 \\ 
  \textsc{Energy}                 & 2.3  & 38.6 \\ 
  \textsc{Utilities}              & 2.2  & 52.2 \\ 
  \textsc{Materials}              & 1.5  & 40.7 \\ 
   \bottomrule
\end{tabular}
}
{\hspace{-0.5cm} \emph{Note:} Business sectors are categorized according to the Global Industry Classification Standard (GICS). Startups can belong to multiple business sectors.}
\end{table}

\subsection{Baselines}
\label{sec:Baselines}

We compare our $k$NN-ICL against a set of supervised and in-context learning baselines that differ in their learning paradigm and access to historical data. The baselines are designed to isolate (i) the role of inference-time reasoning versus model training and (ii) the contribution of retrieval-augmented few-shot learning. Importantly, our experimental design thereby distinguishes between two evaluation objectives. First, when comparing against supervised ML models, we control for data access by providing all models with the \textit{same retrieved examples} (i.e., there is no advantage for $k$NN-ICL by having access to a larger dataset or by choosing a better set of data points). Second, when comparing different ICL variants, we allow each method to retrieve examples from the full historical pool in order to directly evaluate the quality of the retrieval mechanism itself. All baseline variants use the same data, the same LLM, the same prompt design, and the same evaluation protocol to ensure fair comparisons. Accordingly, any performance improvements observed for the proposed approach can be attributed to the $k$NN-ICL framework rather than differences in data or evaluation settings.

\vspace{0.3cm}

\noindent
\textbf{\underline{Baselines from supervised ML.}}  
We follow prior literature \citep[e.g.,][]{Arroyo.2019,Krishna.2016,Zbikowski.2021} and include traditional supervised learning models trained on the same structured and unstructured data $(\,\mathbf{x}^{\text{SD}}_i,\mathbf{x}^{\text{TD}}_i\,)$ as used in our $k$NN-ICL. Thereby, we aim to show that supervised models are often not suitable for operating in data-scarce environments. Crucially, to ensure a fair comparison with $k$NN-ICL, all supervised ML models are trained on the same set of retrieved startups that form the in-context examples for $k$NN-ICL. This design choice removes differences in information access as a confounding factor. In particular, it ensures that any performance differences between supervised ML and $k$NN-ICL are not driven by superior candidate selection or access to a larger historical pool, but instead reflect differences in how the models utilize the \textit{same limited evidence}). In this sense, the comparison isolates the effect of inference-time reasoning via ICL versus parameter training via supervised learning. Later, we compare the following supervised ML  baselines:
\begin{itemize}[leftmargin=*]
\item \emph{Regularized logistic regression.}  Logistic regression is a standard baseline in startup prediction research \citep{Choi.2024,Sharchilev.2018}, which models the log-odds of success as a linear function of the covariates. Due to the parsimonious structure and strong regularization, this model tends to remain robust in data-scarce environments.
\item \emph{Random forest.}  Random forests have been used in prior work on startup success prediction \citep{RazaghzadehBidgoli.2024}. By aggregating many decorrelated decision trees trained on bootstrapped samples, random forest can capture nonlinear relationships but are often still robust even small sample size setting such as from VC practice. 
\item \emph{XGBoost.}  
XGBoost represents a state-of-the-art gradient boosting method frequently applied in tabular prediction problems, including startup success prediction \citep{Thirupathi.2022}. Similar to random forest, XGBoost is often very robust to overfitting and thus presents a natural baseline for our VC setting.

\end{itemize}
We also experimented with neural networks trained on the same  inputs, that is, using the same shot as in our main setup (10, 20, 30, and 50 training examples per prediction). However, across small sample sizes, neural networks did not yield a competitive performance. This is expected because neural networks typically require substantially more labeled data to learn reliable parameters. With only 10--50 examples, neural networks are prone to overfitting and high variance even with regularization, so we omit them from the main benchmark.

To ensure fair comparison, we train the standard ML baselines using the same shots as in the $k$NN support examples. All standard ML baselines are implemented using standard Python libraries (scikit-learn and XGBoost). Consistent with the data-scarce VC setting, we adopt standard, widely used baseline configurations and limit hyperparameter tuning to avoid overfitting on small training sets. For the regularized logistic regression, we use \texttt{lbfgs} with $\ell_2$ regularization, \texttt{class\_weight=balanced}, \texttt{max\_iter=1000}, and the default regularization strength \texttt{C=1.0}. For the random forest, we fit a \texttt{RandomForestClassifier} with 300 trees, \texttt{class\_weight=balanced}, \texttt{min\_samples\_leaf=2}, and \texttt{n\_jobs=-1}. For the XGBoost baseline, we fit an \texttt{XGBClassifier} with 400 trees, a maximum depth of 6, a learning rate of 0.05, subsampling and column-subsampling rates of 0.8, and $\ell_2$ regularization (\texttt{reg\_lambda=1.0}).

\vspace{0.3cm}

\noindent
\underline{\textbf{ICL variants.}} To understand the role of ICL, we evaluate several variants of our ICL approach that differ in how shots are selected. These ablation studies allow us to determine whether our proposed $k$NN-ICL framework outperforms simpler ICL approaches. In contrast to the supervised ML comparison above, these variants are allowed to retrieve examples from the full historical pool. This evaluation protocol intentionally evaluates the effectiveness of different retrieval strategies rather than holding retrieval fixed.
\begin{itemize}[leftmargin=*]
\item \emph{Zero-shot LLM.} As a lower-bound baseline, we consider a zero-shot configuration in which no labeled examples are provided in the prompt. The model is asked to predict startup success based solely on the target startup’s description and structured attributes, allowing us to assess the predictive capability of the underlying LLM in the absence of in-context demonstrations. The zero-shot prompt is identical to the prompt in the $k$NN-ICL framework, except that no in-context examples are provided and that this part of the prompt is just omitted.
\item \emph{Vanilla ICL.} As a reference point for evaluating the benefit of data-driven example selection, we also consider a standard ICL setup in which shots are selected \emph{at random}. This baseline allows us to isolate the contribution of the retrieval mechanism by comparing $k$NN-ICL against an ICL variant that uses the same model and prompt structure but does not exploit similarity-based selection. In doing so, we can assess whether targeted example retrieval provides an improvement over vanilla random sampling. The prompt is otherwise identical to that of $k$NN-ICL but the choice and order of the demonstrations is different.
\item \emph{Sector-focused ICL.}  To assess whether domain-specific context improves prediction, we evaluate a simple baseline that selected half of the shots from startups operating in the same sector as the target company. The remaining shots are sampled randomly from the rest of the training pool. This configuration reflects the intuition that sector-related peer startups may provide more relevant reference cases, while preserving diversity across other industries. This presented a heuristic approach to strategically select startup examples, but without matching on the entire company profile. We compute sector overlap using the one-hot encoded sector indicators in our dataset. For each target startup, we identify a sector pool by selecting all training startups that share at least one active sector label with the target, i.e., those for which the element-wise overlap across sector columns is non-zero. If a startup has no recorded sector information, the sector pool is left empty and the remaining examples are sampled from the general training set.
\end{itemize}

\subsection{Evaluation Metrics}
\label{sec:evaluation_metrics}

To assess the performance of our $k$NN-ICL approach in startup success prediction, we follow previous research \citep{Maarouf.2025} and report four key classification metrics: balanced accuracy, precision, recall, and $F_1$-score. Our primary evaluation metric is the balanced accuracy (Bal-Acc), which is designed to address the unequal class distribution of startup success data, where unsuccessful startups are more common than successful startups. Unlike the `standard' accuracy, the balanced accuracy equally weights both classes by averaging their true positive rates, ensuring that performance reflects a model’s ability to generalize across both successful and unsuccessful startups. Further, we additionally report precision, recall, and the $F_1$-score. Precision captures the accuracy of `successful' predictions, recall measures the proportion of truly successful startups correctly identified, and the $F_1$-score combines precision and recall using the harmonic mean into a single metric that is informative under class imbalance. All metrics are computed on held-out test sets from each train-test split in a five-repeated stratified shuffle-split procedure based on the model's outputs (i.e., \texttt{"SUCCESS"} or \texttt{"FAILURE"}). We report the mean and standard deviation across test-train splits to evaluate robustness under different random partitions.

Memorization is carefully addressed by swapping each startup’s name in the textual description with a generic placeholder \emph{(''this company'')}. This ensures that the prompt does not expose the real company name and prevents the LLM from leveraging any prior knowledge to that name during pretraining. 

\subsection{Experimental Details}

Our numerical experiments follow established best practices in ML \citep{Feuerriegel.2025}. To assess out-of-sample performance, we employ five-repeated stratified shuffle splits that assigns 80\,\% of the startups to the training set and 20\,\% to the test set in each split. For each variant, predictions were generated on held-out startups in every split, and the resulting scores were summarized as mean and standard deviation across all runs. Stratification ensures that the class balance between successful and unsuccessful startups is preserved within every split. The procedure is repeated five times to obtain robust estimates of predictive performance across random partitions, thereby quantifying how well the models generalize to startups unseen during training. As stated above, the potential risk of memorization by the LLM is addressed by masking any company names in the text description.

Within each split, a \texttt{StandardScaler} is fit \emph{only} on the training split and applied to train/test to avoid leakage. Predictions are performed with \texttt{Gemini~2.0~Flash} using a low temperature (0.2), and we parse the response case-insensitively and map any string beginning with ``\texttt{SUCCESS}'' to the positive class. Textual self-descriptions are embedded with \texttt{embedding-001} using the \texttt{google.generativeai} Python SDK (version~0.6.1). Embedding calls are LRU-cached and protected by exponential backoff on transient rate limits. We form a fused vector by concatenating the embedding with the scaled numeric features.  All codes are implemented in Python~3.9.12.

\section{Results}
\label{sec:results}

This section reports the empirical results of our $k$NN-ICL framework. We first compare $k$NN-ICL against supervised ML baselines under identical data splits (Section~\ref{sec:comp_supervised}). Next, we examine how performance differs across ICL variants (Section~\ref{sec:icl_variants}) and isolate the contribution of structured versus textual inputs (Section~\ref{sec:text_vs_stc}). Finally, we report prediction performance across business sectors to assess heterogeneity across domains (Section~\ref{sec:sector_perf}). An additional comparison against fine-tuned LLMs is in Appendix~\ref{appendix:comp_llm}.

\subsection{Comparison of Our $k$NN-ICL Against Supervised ML}
\label{sec:comp_supervised}

We now evaluate the performance of our $k$NN-ICL for predicting startup success (see Table~\ref{tab:icl_log}). Consistent with common practice in ML, all results are based on out-of-sample predictions for startups that were not part of the training set (e.g., not used as the demonstrations in the prompt) to ensure that the performance is assessed on unseen cases. To account for potential variation across different sample partitions, we repeat the analysis using multiple train–test splits and report the mean and standard deviation of all performance metrics across these repetitions. 

Importantly, we carefully addressed the risk of memorization in LLMs as follows. We removed company names, so that no company-identifying information is exposed in the prompt, which reduces the risk of data leakage. Moreover, to ensure a fair comparison between inference-time ICL and supervised learning, all supervised ML baselines are trained and evaluated on exactly the same retrieved startup examples that are provided as in-context demonstrations to $k$NN-ICL. This choice of the evaluation protocol holds the information set constant across methods and ensures that differences in performance are attributable \textit{only} to the modeling paradigm—training-based learning versus inference-time reasoning---rather than differences in candidate selection or access to a larger historical pool. Accordingly, any observed performance differences reflect the relative effectiveness of the prediction approach and not differences in data composition.

In Table~\ref{tab:icl_log}, we compare the proposed $k$NN-ICL approach against traditional supervised ML baselines, namely, regularized logistic regression, random forest, and XGBoost. We exclude neural network models because they require large dataset for tuning and, therefore, typically do not perform well on datasets of this size. Our analysis shows the following key patterns: (1)~Across all settings, $k$NN-ICL outperforms regularized logistic regression in both balanced accuracy and $F_{1}$-score. For $k=50$ examples, $k$NN-ICL achieves a balanced accuracy of 0.713 and an $F_{1}$-score of 0.632, compared with 0.631 and 0.488 for logistic regression, which correspond to an improvement of 8.2 percentage points in balanced accuracy and 14.4 points in $F_{1}$-score. The performance gain is robust for all key metrics (in a few cases, logistic regression shows slightly higher precision, but differences remain within one standard deviation). (2)~Even when compared to random forest and XGBoost, which reach balanced accuracies of 0.634 and 0.618, respectively, $k$NN-ICL remains superior across all metrics. (3)~The performance gap between $k$NN-ICL and supervised baselines widens as the number of examples increases. This highlights the ability of in-context reasoning to scale with additional demonstrations. At the same time, the performance saturates at around $k=50$, suggesting that there is a natural trade-off:  larger support sets provide more information but may also introduce less relevant examples, which thus hampers performance.

\begin{table}[H]
\TABLE
{Prediction performance compared to supervised ML. 
\label{tab:icl_log}}
{
\OneAndAHalfSpacedXI
\footnotesize
\renewcommand{\arraystretch}{1.15}
\sisetup{
  round-mode=places,
  round-precision=3,
  detect-weight,
  mode=text
}

\begin{tabular}{
l
l
>{\centering\arraybackslash}p{1.55cm}
>{\centering\arraybackslash}p{1.55cm}
>{\centering\arraybackslash}p{1.55cm}
>{\centering\arraybackslash}p{1.55cm}
}
  \toprule
  {\textbf{Shots ($k$)}} & {\textbf{Variant}} 
  & {\thead{\textbf{Balanced} \\ \textbf{accuracy}}} 
  & {\textbf{$\boldsymbol{F_1}$-score}} 
  & {\textbf{Precision}} 
  & {\textbf{Recall}} \\
  \midrule

  \multirow{4}{*}{10}
        & Regularized logistic regression 
      & 0.647 (0.015)
      & 0.537 (0.020)
      & \textbf{0.557} (0.021)
      & 0.518 (0.020) \\

        & Random forest
      & 0.616 (0.012)
      & 0.504 (0.015)
      & 0.501 (0.016)
      & 0.509 (0.018) \\

        & XGBoost
      & 0.537 (0.010)
      & 0.441 (0.011)
      & 0.389 (0.010)
      & 0.509 (0.018) \\

     & $k$NN-ICL         
      & \textbf{0.675} (0.017)
      & \textbf{0.595} (0.020)
      & 0.524 (0.018)
      & \textbf{0.690} (0.041) \\

  \midrule

   \multirow{4}{*}{20}
        & Regularized logistic regression 
      & 0.652 (0.021)
      & 0.536 (0.027)
      & 0.589 (0.039)
      & 0.492 (0.020) \\

        & Random forest
      & 0.617 (0.009)
      & 0.498 (0.011)
      & 0.512 (0.017)
      & 0.485 (0.009) \\

        & XGBoost
      & 0.579 (0.011)
      & 0.468 (0.014)
      & 0.443 (0.012)
      & 0.496 (0.019) \\

    & $k$NN-ICL         
      & \textbf{0.697} (0.012)
      & \textbf{0.605} (0.018)
      & \textbf{0.617} (0.010)
      & \textbf{0.594} (0.029) \\
  \midrule

  \multirow{4}{*}{30}
        & Regularized logistic regression 
      & 0.657 (0.021)
      & 0.537 (0.029)
      & \textbf{0.613} (0.042)
      & 0.478 (0.022) \\

        & Random forest
      & 0.619 (0.015)
      & 0.499 (0.022)
      & 0.517 (0.021)
      & 0.482 (0.025) \\

        & XGBoost
      & 0.603 (0.009)
      & 0.492 (0.010)
      & 0.478 (0.014)
      & 0.507 (0.014) \\

    & $k$NN-ICL         
      & \textbf{0.704} (0.021) 
      & \textbf{0.623} (0.024)
      & 0.576 (0.028)
      & \textbf{0.680} (0.032) \\
  \midrule

  \multirow{4}{*}{50}
        & Regularized logistic regression 
      & 0.632 (0.019)
      & 0.488 (0.028)
      & \textbf{0.603} (0.041)
      & 0.411 (0.025) \\

        & Random forest
      & 0.631 (0.020)
      & 0.513 (0.033)
      & 0.537 (0.024)
      & 0.492 (0.047) \\

        & XGBoost
      & 0.624 (0.021)
      & 0.507 (0.029)
      & 0.521 (0.030)
      & 0.494 (0.031) \\

    & $k$NN-ICL         
      & \textbf{0.713} (0.019) 
      & \textbf{0.632} (0.026)
      & 0.596 (0.011)
      & \textbf{0.673} (0.047) \\

  \bottomrule
\end{tabular}
}
{\hspace{-0.5cm} \emph{Note:} Reported are mean values with standard deviations in parentheses. Best performance per metric and configuration highlighted in bold.}
\end{table}

\subsection{Performance Comparison Across ICL Variants}
\label{sec:icl_variants}

Table~\ref{tab:icl_all} compares the prediction performance across different ICL variants. In contrast to the comparison with supervised ML baselines, all ICL variants are allowed to retrieve in-context examples from the full historical pool of startups (more precisely, from the training set of the corresponding split). This choice of the evaluation protocol intentionally evaluates the effectiveness of different retrieval strategies from example selection.

\begin{table}[t]
\TABLE
{Prediction performance across ICL variants by number of examples. 
\label{tab:icl_all}}
{
\OneAndAHalfSpacedXI
\footnotesize
\renewcommand{\arraystretch}{1.15}
\sisetup{round-mode=places,round-precision=3,detect-weight,mode=text}

\begin{tabular}{
l
l
>{\centering\arraybackslash}p{1.55cm}
>{\centering\arraybackslash}p{1.55cm}
>{\centering\arraybackslash}p{1.55cm}
>{\centering\arraybackslash}p{1.55cm}
}
  \toprule
  {\textbf{Shots ($k$)}} & {\textbf{Variant}} 
  & {\thead{\textbf{Balanced} \\ \textbf{accuracy}}} 
  & {\textbf{$\boldsymbol{F_1}$-score}} 
  & {\textbf{Precision}} 
  & {\textbf{Recall}} \\
  \midrule

  \multirow{1}{*}{---}
        & Zero-shot LLM
      & 0.652 (0.019)
      & 0.551 (0.026)
      & 0.543 (0.019)
      & 0.561 (0.036) \\
  \midrule

  \multirow{3}{*}{10}
        & Vanilla ICL
      & 0.673 (0.016)
      & 0.595 (0.018)
      & 0.520 (0.015)
      & \textbf{0.694} (0.027) \\
\addlinespace
        & Sector-focused ICL
      & 0.673 (0.015)
      & 0.585 (0.022)
      & \textbf{0.547} (0.007)
      & 0.630 (0.045) \\
\addlinespace
        & $k$NN-ICL         
      & \textbf{0.675} (0.017)
      & \textbf{0.595} (0.020)
      & 0.524 (0.018)
      & 0.690 (0.041) \\

  \midrule

  \multirow{3}{*}{20}
        & Vanilla ICL
      & 0.693 (0.019)
      & \textbf{0.613} (0.022)
      & 0.552 (0.019)
      & \textbf{0.690} (0.035) \\
\addlinespace
        & Sector-focused ICL
      & 0.691 (0.010)
      & 0.609 (0.012)
      & 0.559 (0.012)
      & 0.670 (0.024) \\
\addlinespace
        & $k$NN-ICL
      & \textbf{0.697} (0.012)
      & 0.605 (0.018)
      & \textbf{0.617} (0.010)
      & 0.594 (0.029) \\
  \midrule
  \multirow{3}{*}{30}
        & Vanilla ICL
      & 0.686 (0.009)
      & 0.605 (0.011)
      & 0.546 (0.010)
      & \textbf{0.680} (0.029) \\
\addlinespace
        & Sector-focused ICL
      & 0.690 (0.014)
      & 0.599 (0.021)
      & \textbf{0.593} (0.008)
      & 0.606 (0.045) \\
\addlinespace
        & $k$NN-ICL         
      & \textbf{0.704} (0.021) 
      & \textbf{0.623} (0.024)
      & 0.576 (0.028)
      & \textbf{0.680} (0.032) \\

  \midrule
  \multirow{3}{*}{50}
        & Vanilla ICL
      & 0.702 (0.013)
      & 0.620 (0.018)
      & 0.577 (0.011)
      & 0.670 (0.035) \\
\addlinespace
        & Sector-focused ICL
      & 0.696 (0.016)
      & 0.604 (0.022)
      & \textbf{0.615} (0.017)
      & 0.594 (0.042) \\
\addlinespace
    & $k$NN-ICL         
      & \textbf{0.713} (0.019) 
      & \textbf{0.632} (0.026)
      & 0.596 (0.011)
      & \textbf{0.673} (0.047) \\

  \bottomrule
\end{tabular}
}
{\hspace{-0.5cm}\emph{Note:} Reported are mean values with standard deviations in parentheses. Best performance per metric and configuration highlighted in bold.}
\end{table}

We make the following observations. (1)~Vanilla ICL provides a strong baseline, reaching, for example, a balanced accuracy of 0.702 at $k=50$ shots. Overall, the performance improves with larger support sets but only moderately. (2)~The vanilla ICL outperforms the zero-shot LLM, implying that the examples in the prompt add predictive power. Still, the zero-shot LLM prompt has an above-random-guess performance, which is expected, simply because the LLM can assess the quality of the business model from the textual description. This mirrors previous research findings that LLMs often work as zero-shot prediction engines for a variety of classification tasks \citep{Feuerriegel.2025}. (3)~The sector-focused ICL leads to mixed results (e.g., the sector-focused ICL is better than the vanilla ICL for $k=30$, but the vanilla ICL is better for $k=50$). (4)~Overall, our $k$NN-ICL performs best and consistently outperforms the other ICL variants. For example, compared to the vanilla ICL, our $k$NN-ICL improves balanced accuracy by roughly 1.7 percentage points and $F_{1}$ by 1.3 points at 50 shots. Because all ICL variants share the same LLM, prompt structure, and number of shots, the performance gains can be attributed directly to the retrieval mechanism rather than to differences in the model or information access. In sum, the findings demonstrate that the $k$NN-ICL framework outperforms other ICL variants in early-stage startup prediction.

\subsection{Performance of Structured vs. Textual Data}
\label{sec:text_vs_stc}

To assess how different information sources contribute to ICL, we compare the prediction performance of our $k$NN-ICL that uses only structured variables (such as founding year, team size, and sector indicators) versus only textual descriptions. This comparison also evaluates how well our ICL approach generalizes to heterogeneous data types, given that LLMs are known to handle text naturally but may struggle with numerical representations or mixed feature formats. By isolating each information source, we can examine the extent to which structured and unstructured data independently support few-shot reasoning.

\begin{table}[t]
\TABLE
{Prediction performance of $k$NN-ICL for different data inputs (i.e., structured vs. textual information about startups). 
\label{tab:icl_struct_text}}
{
\OneAndAHalfSpacedXI
\footnotesize
\renewcommand{\arraystretch}{1.15}
\sisetup{round-mode=places,round-precision=3,detect-weight,mode=text}

\begin{tabular}{
l
l
>{\centering\arraybackslash}p{1.55cm}
>{\centering\arraybackslash}p{1.55cm}
>{\centering\arraybackslash}p{1.55cm}
>{\centering\arraybackslash}p{1.55cm}
}
  \toprule
  {\textbf{Shots ($k$)}} & {\textbf{Variant}} 
  & {\thead{\textbf{Balanced} \\ \textbf{accuracy}}} 
  & {\textbf{$\boldsymbol{F_1}$-score}} 
  & {\textbf{Precision}} 
  & {\textbf{Recall}} \\
  \midrule

  \multirow{3}{*}{10}
        & Structured-only
      & 0.639 (0.013) & 0.543 (0.017) & 0.511 (0.015) & 0.579 (0.028) \\
        & Text-only
      & 0.642 (0.014) & 0.524 (0.023) & \textbf{0.562} (0.015) & 0.492 (0.031) \\
    & Structured + text           
      & \textbf{0.675} (0.017)
      & \textbf{0.595} (0.020)
      & 0.524 (0.018)
      & \textbf{0.690} (0.041) \\
  \midrule

   \multirow{3}{*}{20}
        & Structured-only
      & 0.636 (0.013) 
      & 0.538 (0.017) 
      & 0.510 (0.014) 
      & 0.570 (0.020) \\
        & Text-only
      & 0.645 (0.012) 
      & 0.545 (0.014) 
      & 0.528 (0.018) 
      & 0.564 (0.020) \\
    & Structured + text
      & \textbf{0.697} (0.012)
      & \textbf{0.605} (0.018)
      & \textbf{0.617} (0.010)
      & \textbf{0.594} (0.029) \\
  \midrule

  \multirow{3}{*}{30}
    & Structured-only     
      & 0.629 (0.018) & 0.530 (0.025) & 0.499 (0.018) & 0.565 (0.036) \\
        & Text-only
      & 0.648 (0.017) & 0.543 (0.026) & 0.542 (0.016) & 0.546 (0.039) \\
    & Structured + text    
      & \textbf{0.704} (0.021) 
      & \textbf{0.623} (0.024)
      & \textbf{0.576} (0.028)
      & \textbf{0.680} (0.032) \\
  \midrule

  \multirow{3}{*}{50}
    & Structured-only     
      & 0.606 (0.007) & 0.505 (0.009) & 0.472 (0.008) & 0.544 (0.019) \\
        & Text-only
      & 0.650 (0.005) & 0.546 (0.006) & 0.550 (0.015) & 0.542 (0.018) \\
    & Structured + text    
      & \textbf{0.713} (0.019) 
      & \textbf{0.632} (0.026)
      & \textbf{0.596} (0.011)
      & \textbf{0.673} (0.047) \\
  \bottomrule
\end{tabular}
}
{\hspace{-0.5cm}\emph{Note:} Mean performance shown with standard deviations in parentheses. Best value per metric and configuration highlighted in bold.}
\end{table}

The results are in Table~\ref{tab:icl_struct_text}. (1)~The structured-only $k$NN-ICL performs slightly better at very small support sizes. For $k=10$ shots, structured-only yields a marginally higher $F_1$-score (0.543 vs. 0.524), reflecting that with very limited examples, low-dimensional numeric features can be easier for the model to exploit than sparse textual cues. (2)~As the number of demonstrations $k$ increases, textual descriptions gain in predictive power. The advantage is especially pronounced for $k=30$ shots (0.648 vs. 0.629) and $k=50$ shots (0.650 vs. 0.606). This suggests, consistent with prior research \citep{Maarouf.2025}, that textual descriptions about startups provide powerful predictive signals that LLMs can effectively leverage. (3)~Combining structured and textual data yields the strongest performance. While each information source contributes differently depending on the number of shots, the full $k$NN-ICL approach, which integrates both, consistently outperforms both the structured-only and text-only variants. This suggests that the structured variables provide complementary signals to the semantic information in textual descriptions and that LLMs can effectively synthesize these heterogeneous inputs.

\subsection{Prediction Performance Across Business Sectors}
\label{sec:sector_perf}

To assess whether predictive accuracy varies across industries, we evaluate the $k$NN-ICL framework separately for each GICS sector. Such sector-level analysis is important because both the amount of available data and the homogeneity of business models differ substantially across industries, which may influence how effectively ICL generalizes. The results with a breakdown by GICS sectors are reported in Table~ \ref{tbl:Sectors_performance}. The \textsc{Utilities} sector attains the highest balanced accuracy (76.9\%), followed by \textsc{Financials} (75.6\%). In contrast, a lower prediction performance is observed for \textsc{Energy} (66.5\%) and \textsc{Materials} (67.6\%). Yet, both of the sectors are among the smallest cohorts in the dataset and thus are likely to be affected by sampling variability. In summary, the $k$NN-ICL approach delivers a robust prediction performance across different industries, with only minor fluctuations driven primarily by data scarcity rather than by sector-specific model limitations.

\begin{table}[H]
\TABLE
{Prediction performance across business sectors.\label{tbl:Sectors_performance}}
{
\OneAndAHalfSpacedXI
\tiny
\sisetup{detect-weight,mode=text}
\sisetup{round-mode=places,round-precision=2,table-column-width=1.3cm}
\begin{tabular}{l
                S[table-format=2.2]
                S[table-format=2.2]
                S[table-format=2.2]
                S[table-format=2.2]
                S[table-format=2.2]}
  \toprule
  {\textbf{Business sector}} 
    & {\thead{\tiny\textbf{Relative} \\ \tiny\textbf{freq. (in\,\%)}}} 
    & {\thead{\tiny\textbf{Balanced} \\ \tiny\textbf{accuracy}}} 
    & {\textbf{$\boldsymbol{F_1}$-score}} 
    & {\textbf{Precision}} 
    & {\textbf{Recall}} \\
  \midrule
  \textsc{Utilities} & 1.09 & 76.93 & 77.42 & 78.26 & 76.60 \\
  \textsc{Financials} & 4.26 & 75.65 & 69.93 & 65.79 & 74.63 \\
  \textsc{Real Estate} & 2.99 & 71.37 & 62.72 & 67.09 & 58.89 \\
  \textsc{Consumer Discretionary} & 12.49 & 70.81 & 60.89 & 58.29 & 63.74 \\
  \textsc{Communication Services} & 23.30 & 70.69 & 60.67 & 57.38 & 64.35 \\
  \textsc{Information Technology} & 28.08 & 70.65 & 65.26 & 60.60 & 70.70 \\
  \textsc{Industrials} & 15.09 & 70.29 & 68.34 & 65.07 & 71.96 \\
  \textsc{Consumer Staples} & 2.66 & 69.93 & 69.12 & 70.09 & 68.18 \\
  \textsc{Materials} & 0.81 & 67.59 & 56.52 & 54.17 & 59.09 \\
  \textsc{Health Care} & 8.05 & 67.28 & 68.83 & 65.42 & 72.62 \\
  \textsc{Energy} & 1.18 & 66.45 & 53.12 & 48.57 & 58.62 \\
  \bottomrule
\end{tabular}
}
{
\hspace{-0.5cm} \emph{Note:} Metrics reported in \%. Relative frequencies report the share of test predictions per sector
summed across all five stratified shuffle-split repetitions. Model: $k$NN-ICL (50 examples per prediction).
}
\end{table}

\subsection{Sensitivity to the Fusion Scaling Parameter $\alpha$}
\label{sec:scaling}

As a robustness check, we vary the fusion scaling parameter $\alpha$, which controls the relative contribution of the structured feature block $\mathbf{x}_i$ versus the textual embedding $\mathbf{e}_i$ in cosine-similarity retrieval (Table~\ref{tab:alpha_sensitivity}). Across $\alpha \in \{0.3,0.4,0.5,0.6,0.7\}$ in the 50-shot setting, predictive performance remains stable, with balanced accuracy ranging from 0.705 to 0.709 and $F_1$ from 0.614 to 0.620. Increasing $\alpha$ slightly improves recall (from 0.593 to 0.606) and yields the best overall trade-off at $\alpha=0.5$, while precision stays essentially unchanged (0.637 versus 0.638). Overall, these results suggest that retrieval quality and downstream ICL predictions are not overly sensitive to moderate changes in $\alpha$, supporting the robustness of our $k$NN-ICL design.

\begin{table}[H]
\TABLE
{Sensitivity of $k$NN-ICL performance to the fusion scaling parameter $\alpha$ (50-shot setting).%
\label{tab:alpha_sensitivity}}
{
\OneAndAHalfSpacedXI
\footnotesize
\renewcommand{\arraystretch}{1.15}
\sisetup{round-mode=places,round-precision=3,detect-weight,mode=text}

\begin{tabular}{
l
>{\centering\arraybackslash}p{1.55cm}
>{\centering\arraybackslash}p{1.55cm}
>{\centering\arraybackslash}p{1.55cm}
>{\centering\arraybackslash}p{1.55cm}
}
\toprule
{\textbf{$\alpha$}} 
& {\thead{\textbf{Balanced} \\ \textbf{accuracy}}} 
& {\textbf{$\boldsymbol{F_{1}}$-score}} 
& {\textbf{Precision}} 
& {\textbf{Recall}} \\
\midrule

0.3 
& 0.705 (0.015)
& 0.614 (0.022)
& \textbf{0.638} (0.014)
& 0.593 (0.033) \\

0.4 
& 0.706 (0.013)
& 0.616 (0.020)
& 0.637 (0.019)
& 0.598 (0.040) \\

0.5 
& \textbf{0.709} (0.014)
& \textbf{0.620} (0.019)
& 0.637 (0.018)
& \textbf{0.606} (0.033) \\

0.6
& 0.701 (0.014)
& 0.609 (0.023)
& 0.632 (0.011)
& 0.589 (0.039) \\

0.7
& 0.697 (0.009)
& 0.604 (0.014)
& 0.627 (0.013)
& 0.585 (0.029) \\

\bottomrule
\end{tabular}
}
{\hspace{-0.5cm} \emph{Note:} Reported are mean values with standard deviations in parentheses across the five repeated stratified shuffle-split runs. The parameter $\alpha$ rescales the structured block prior to concatenation, controlling the relative influence of structured versus textual information in cosine-similarity retrieval.}
\end{table}

\section{Discussion}
\label{sec:disscusion}

\subsection{Managerial Implications}
\label{sec:managerial_implications}

% findings

Our results highlight the practical value of using ICL for predicting startup success. The results show that our proposed framework can generate accurate predictions of startup success without any model training or access to large training datasets. Specifically, our $k$NN-ICL framework achieves a balanced accuracy of 71.3\% and an $F_1$-score of 0.63. Importantly, it outperforms both supervised ML baselines (which are fine-tuned) and a vanilla ICL (which uses few-shot learning based on randomly selected examples). Moreover, the performance gains are robust across different values of $k$ and across varying numbers of available in-context examples. Hence, our method is of direct practical relevance to investors and organizations seeking to identify promising startups, especially in the early stage when data is scarce.

% takeaway: small sample sizes => ICL 

Our work directly addresses a key challenge in managerial practice: many organizations operate in small-data environments where training ML models is difficult, costly, or infeasible. In particular, access to large labeled datasets is often infeasible in many business applications, which renders traditional supervised ML approaches impractical. A potential remedy is offered by ICL \citep{Brown.2020,Dong.2024}, which allows organizations to draw on a small set of relevant past cases to make predictions at \textit{inference time} (i.e., without any model training or feature engineering). Yet, the operational value of ICL for business decisions has remained largely unexplored. Our findings show that ICL provides an easy-to-implement way to leverage pretrained LLMs for predictions in settings with inherent data scarcity. Notably, even very small support sets (e.g., $k=10$) already outperform fine-tuned supervised ML models, despite when the latter have access to considerably larger training datasets (e.g., XGBoost for $k=50$). These results highlight that few-shot ICL can deliver strong predictive performance in data-scarce settings.

% interpretability

Even though LLMs are not inherently interpretable, our $k$NN-ICL approach offers a practical degree of transparency by making the retrieved example set for predictions explicit. Managers can inspect which past startups the model draws upon, allowing them to reflect on how these cases may inform the prediction. In particular, analysts can benchmark new startups against similar ventures with known outcomes to assess investment decisions. In this sense, ICL provides a form of case-based interpretability that aligns with familiar tools such as scorecards or comparison sets, thereby helping decision makers understand why a prediction may be plausible. Nevertheless, deeper interpretability remains an open area for future research.

\subsection{Methodological Implications}
\label{sec:methodological_implications}

Our work contributes to the growing research that leverages LLMs for operations research and business analytics \citep[e.g.,][]{Maarouf.2025, Wu.2025}. The ICL paradigm represents an important methodological shift in practice: rather than learning patterns through parameter optimization, the model reasons directly from examples provided in the prompt. Our experiments show that this approach performs well even when labeled data are scarce, thereby offering a viable alternative to traditional supervised pipelines. Here, we add to this emerging literature by demonstrating how ICL can be operationalized for prediction tasks in business analytics with minimal data and by developing a retrieval-based mechanism that systematically selects the most informative examples for inference. Additionally, to address memorization, we remove company names from startup textual descriptions to reduce the risk that predictions are influenced by prior knowledge acquired during model training.

% method

Methodologically, our results highlight several important trade-offs in applying ICL to operational prediction tasks. First, the relevance of the selected examples is crucial: ICL performs substantially better when the model is prompted with startups that closely resemble the target case. For this, we develop our $k$NN-ICL framework, which automatically retrieves comparable startups, and our experiments show that such data-driven retrieval improves accuracy relative to vanilla ICL with randomly chosen examples. Second, the number of demonstrations ($k$) plays an important role in few-shot learning. Our findings reflect earlier observations \citep[e.g.,][]{Schulhoff.2025}, where larger values of $k$ provide the model with richer contextual information, but excessively large sets can dilute relevance and hinder performance; notably, even small values of $k=10$ or $k=30$ already yield a strong predictive performance, while the performance gains saturate at around $k=50$. Still, future LLMs with better reasoning and longer context windows may also be able to handle a larger number of examples. Third, our ICL approach naturally integrates structured and unstructured data, allowing companies to accommodate both numerical signals and textual descriptions to make predictions in a unified framework. 

% future work

Looking ahead, we see substantial potential for applying ICL in a wide range of operational settings characterized by scarce or heterogeneous data. Many prediction problems in practice resemble our task above: organizations must make decisions with only a handful of historical examples and limited structured information. We see particularly promising applications in business failure forecasting \citep[e.g.,][]{Borchert.2023}, credit risk evaluation \citep[e.g.,][]{Gunnarsson.2021,Kriebel.2022}, and other domains where textual descriptions, expert notes, or unstructured documents play a central role. Overall, we expect that retrieval-augmented few-shot learning will become an increasingly valuable tool for operational decision-making across industries.

\subsection{Limitations and Future Research}
\label{sec:limitations}

Our study has several limitations that suggest directions for future research. Our empirical setup was carefully curated to ensure no information leakage. Although Crunchbase is larger and more comprehensive than the datasets typically available to VC firms, and although we rely on publicly accessible data for reproducibility, our benchmarking was intentionally designed to reflect the smaller and sparser information environment in which early-stage investors operate. Still, VC firms may wish to apply $k$NN-ICL using their own proprietary data, which could include additional structured data or richer unstructured data such as pitch decks or business plans; incorporating such data is straightforward with our framework. In addition, we define startup success as a binary outcome that refers to an acquisition or IPO within a fixed time horizon. While this evaluation metric is consistent with prior literature \citep{Arroyo.2019,Hegde.2021}, it does not capture the broader range of potential startup trajectories, such as sustainable long-term operation, gradual growth, or moderate returns without a formal exit. Future work could explore alternative outcome definitions.

\section{Conclusion}
\label{sec:conclusion}

Only a small share of early-stage startups ultimately succeed, making investment decisions challenging, yet many VC firms have access to only limited historical data. To address this challenge, we propose a few-shot learning framework based on a novel $k$NN-ICL approach, which allows us to predict startup success from a small set of startups from historical databases. Our empirical results show that our approach substantially improves prediction accuracy over a vanilla ICL and standard supervised ML baselines. More broadly, our framework illustrates the operational value of retrieval-augmented, few-shot learning for decision support in data-scarce environments.

\begin{APPENDICES}

\section{Comparison against other LLM approaches}
\label{appendix:comp_llm}

To contextualize the performance of our approach, we further compare against the fused LLM introduced by \citet{Maarouf.2025}. Their fused LLM is trained on 20{,}172 startups from the Crunchbase dataset and combines structured fundamental variables with textual self-descriptions in a fused neural architecture. We evaluated both their fused LLM and our $k$NN-ICL on the same test set to ensure fair comparisons. As expected, the fused LLM benefits from access to the large dataset and achieves a balanced accuracy of 0.743 and an $F_{1}$-score of 0.678. In contrast, our $k$NN-ICL framework operates in a data-scarce setting. Our  $k$NN-ICL framework does \textit{not} undergo any parameter training and, at prediction time, receives only a small set of labeled examples in the prompt. For $k=50$ shots, our $k$NN-ICL achieves a balanced accuracy of 0.713 and an $F_{1}$-score of 0.632. Still, this leads to an interesting finding: even though the fused LLM leverages more than 20{,}000 labeled startup profiles during training, our $k$NN-ICL approach comes close in terms of predictive performance while using only 50 shots per prediction. This shows that our $k$NN-ICL framework performs well, although the fused LLM still achieves higher accuracy.

\begin{table}[H]
\TABLE
{Comparison between the proposed $k$NN-ICL approach and the fused large language model of \cite{Maarouf.2025}.}
{
\OneAndAHalfSpacedXI
\footnotesize
\renewcommand{\arraystretch}{1.15}
\sisetup{round-mode=places,round-precision=3,detect-weight,mode=text}

\begin{tabular}{
l
>{\centering\arraybackslash}p{1.55cm}
>{\centering\arraybackslash}p{1.55cm}
>{\centering\arraybackslash}p{1.55cm}
>{\centering\arraybackslash}p{1.55cm}
>{\centering\arraybackslash}p{2.05cm}
}
\toprule
{\textbf{Model}} 
& {\thead{\textbf{Balanced} \\ \textbf{accuracy}}} 
& {\textbf{$\boldsymbol{F_{1}}$-score}} 
& {\textbf{Precision}} 
& {\textbf{Recall}} 
& {\thead{\textbf{Large-data} \\ \textbf{setting}}} \\
\midrule

Fused LLM (FV + TSD) 
& 0.743 (0.0025) 
& 0.678 (0.0015) 
& 0.598 (0.0179) 
& 0.783 (0.0263)
& \cmark \\

\addlinespace[2pt]
\midrule

$k$NN-ICL (50 shots) 
& 0.713 (0.019) 
& 0.632 (0.026) 
& 0.596 (0.011) 
& 0.673 (0.047)
& \xmark \\

\bottomrule
\end{tabular}
}
{\hspace{-0.5cm} \emph{Note:} Reported are mean values with standard deviations in parentheses. FV = fundamental variables; TSD = textual self-descriptions.}
\end{table}

\end{APPENDICES}

  \newcommand{\dq}{"}
	\renewcommand{\textquotedbl}{"}
\OneAndAHalfSpacedXI
%\SingleSpacedXI
\bibliographystyle{informs2014} % outcomment this and next line in Case 1
{%\OneAndAHalfSpacedXI
\OneAndAHalfSpacedXI
\bibliography{literature} % if more than one, comma separated
}
\OneAndAHalfSpacedXI

\end{document}